\def\year{2020}\relax
\documentclass[letterpaper]{article} 
\usepackage{aaai20}  
\usepackage{times}  
\usepackage{helvet} 
\usepackage{courier}  
\usepackage[hyphens]{url}  
\usepackage{graphicx} 
\usepackage{graphicx}  
\usepackage{amsmath}
\usepackage{multirow}
\usepackage{amsfonts}
\usepackage{textcomp}

\title{REST: Performance Improvement of a Black Box Model \\via RL-based Spatial Transformation }

\author{
Jae Myung Kim\thanks{Equal Contribution}, Hyungjin Kim\footnotemark[1], Chanwoo Park\footnotemark[1], and Jungwoo Lee
\\
Department of Electrical and Computer Engineering, Seoul National University, Seoul, Korea
\\
\{goldkim92, hjkim, cpark\}@cml.snu.ac.kr, junglee@snu.ac.kr
}

\begin{document}

\maketitle


\begin{abstract}

In recent years, deep neural networks (DNN) have become a highly active area of research, and shown remarkable achievements on a variety of computer vision tasks.  DNNs, however, are known to often make overconfident yet incorrect predictions on out-of-distribution samples, which can be a major obstacle to real-world deployments because the training dataset is always limited compared to diverse real-world samples. Thus, it is fundamental to provide guarantees of robustness to the distribution shift between training and test time when we construct DNN models in practice. Moreover, in many cases, the deep learning models are deployed as black boxes and the performance has been already optimized for a training dataset, thus changing the black box itself can lead to performance degradation. We here study the robustness to the geometric transformations in a specific condition where the black-box image classifier is given. We propose an additional learner, \emph{REinforcement Spatial Transform learner (REST)}, that transforms the warped input data into samples regarded as in-distribution by the black-box models. Our work aims to improve the robustness by adding a REST module in front of any black boxes and training only the REST module without retraining the original black box model in an end-to-end manner, i.e. we try to convert the real-world data into training distribution which the performance of the black-box model is best suited for. We use a confidence score that is obtained from the black-box model to determine whether the transformed input is drawn from in-distribution. We empirically show that our method has an advantage in generalization to geometric transformations and sample efficiency.   
\end{abstract}


\begin{figure*}[t]
\centering
\includegraphics[width=0.85\textwidth]{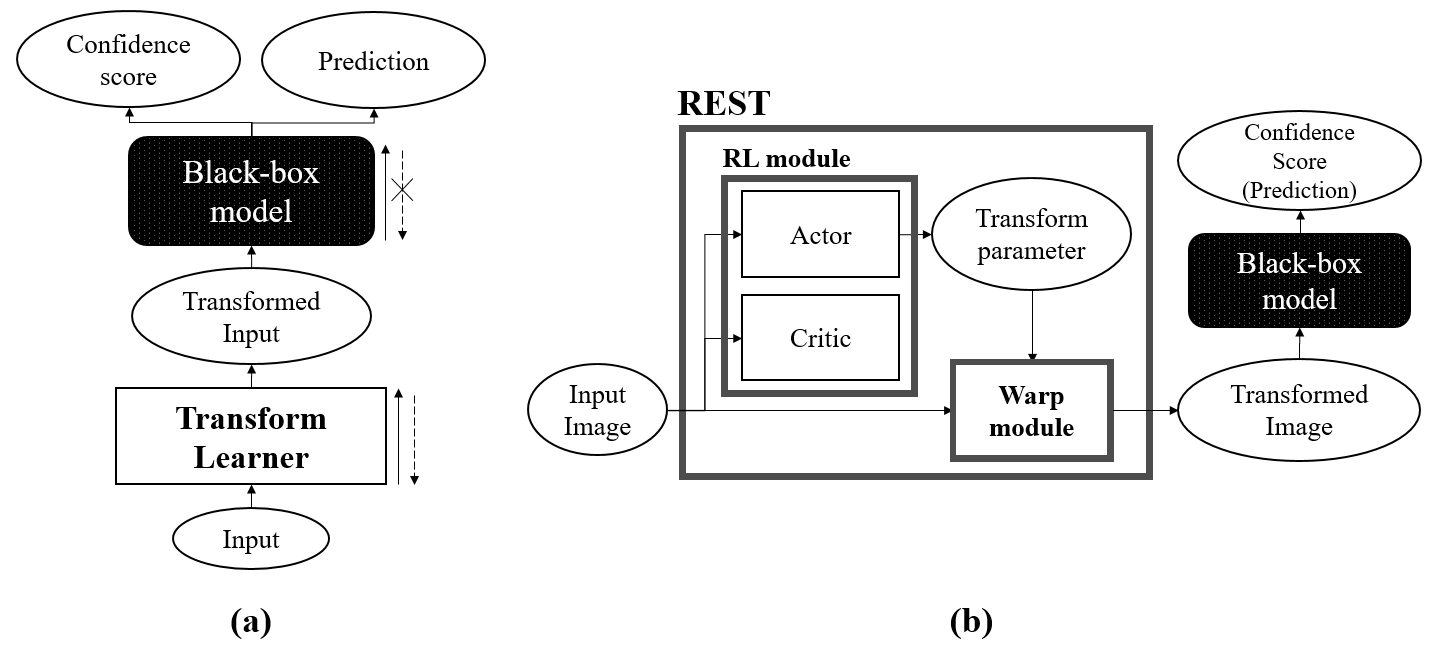} 
\caption{The schematic description. Figure (a) shows the framework of our approach. The white rectangle demonstrates a trainable model and the black represents a black-box model.  Ellipse is an input or an output of the models. Solid lines are forward propagation and dashed lines are backpropagation. Figure (b) shows a more detailed framework where the black-box is an image classifier. The REST learner consists of two modules which are the RL module and the warp module.}
\label{fig1}
\end{figure*}


\section{Introduction}

In the last few years, convolutional neural networks(CNNs) have made great progress in visual recognition tasks such as image classification, object detection, and semantic segmentation\cite{he2016deep,redmon2016you,long2015fully}. Although diverse CNN-based architectures have shown state-of-the-art performance in many computer vision tasks, this result is based on the assumption that training and test data are drawn from the same distribution. However, when the deep learning applications are actually deployed in real-world settings, it is inevitable to deal with real data generated from a different distribution from the training phase, and the performance cannot be guaranteed. As is the case with geometric variations caused by different poses, deformations, and viewpoints, it can be one of the major challenges that deep learning systems face in reality. To address this issue, several studies have been reported in the literature\cite{xu2014scale,marcos2016learning,follmann2018rotationally,you2018prin,worrall2017harmonic}.

Apart from the issue of model performance on real-world data, automated decision-making systems are usually opaque, which means we are not fully accessible to them. The only thing we can observe is the input and the predicted output of black box model. This raises a natural question about the systems. How much should we trust the predictions of the black box system with the given input? Even if we assume there is an indicator that tells us how much we can trust, what actions should we take to obtain reliable predictions if it tells us that the predicted result is highly uncertain? 

For example, let us consider a car number recognition system. The embedded system outputs the predicted number given a license plate. The predictions of the system are usually very accurate given the numbers with typical shapes. However, a crumpled license plate by a car accident (e.g. tilted number 7 written in the plate) would prevent the current system from recognizing the numbers accurately (e.g. predicted as 1). The system would also have high uncertainty in its prediction.

Motivated by the question above, this paper aims to spatially transform the input data to obtain the most reliable predictions by applying the sequence of modified inputs to the black-box model. If test data is not sampled from the same distribution as the data used for training the black-box model, the probability of false prediction increases. To improve accuracy and reliability of predictions, the test input data should be transformed to follow the training data distribution of the black-box model. We refer to the transformation model as a \emph{transform learner}. After the test data is modified by the transform learner, the black-box model maps the transformed data to the output which tells us not only the prediction but also the confidence of the prediction, which is named \emph{confidence score} here. The schematic description is given in Figure 1(a).

The main challenge in our setting is that we do not know what data distribution the black box system has trained from. The prediction and the confidence score are provided by the black-box model only when a particular input is given. We cannot directly train the transform learner by a gradient-based method because the black-box model prevents the gradient flow from reaching to the transform learner. To deal with the problem, we introduce a black-box optimization technique which approximates the gradient of the output by reinforcement learning\cite{jacovi2018neural}. 

Specifically, in this paper, we take a close look at image classification as the black-box problem. We use a pretrained image classifier for the black-box model. Input images are transformed by a transform learner, which we call \emph{REinforcement Spatial Transform learner (REST)}. REST is trained to perform geometric transformation on the input images to increase the confidence score. At inference time, an input image is transformed by REST, followed by predicting the class label via the black-box model.  


\section{Related Work}

\subsection{Invariance to geometric transformations}

There have been many studies to achieve geometric invariance or equivariance in computer vision problems such as classification and segmentation, and we here review four methods: Data augmentation, Spatial Transformer network, Deformable Convolutional Networks, and Capsule network.

\subsubsection{Data augmentation}
Data augmentation (DA)\cite{301simard2003best} is a ubiquitous technique largely used to improves generalization. Although DA is a good starting point for transformation invariance, it is extremely expensive to augment training images with every possible combination of random rotations, shifts, and scales and it is often observed that the models learned with DA are only invariant to known transformations rather than arbitrary changes.

\subsubsection{Spatial Transformer Network}
Spatial transformer networks (STN)\cite{STN_jaderberg2015spatial} is the first work which learns to apply spatial transformation to warp feature maps in an end-to-end fashion. STN consists of three components - localization network, grid generator and sampler. Localization network learns input-dependent transformations and allows the entire network to increase classification accuracy. The drawback of the STN is that it has to be trained with various transformations in the training phase, and we find that STN failed to generalize when unknown, rare transformations are applied to input images.

\subsubsection{Deformable Convolutional Networks}
In \cite{dai2017deformable} the authors argue that it is inherently difficult to handle objects with different scales and shapes in regular CNN because their operations, e.g. convolution kernels, max-pooling, have geometrically fixed patterns. Deformable ConvNets tries to model the dense spatial transformations by learning 2D offsets to the regular sampling grid from target tasks instead of parametric transformations. Deformable ConvNets is then applied to semantic segmentation and object detection to demonstrate its efficiency.

\subsubsection{Capsule network}
Capsules\cite{capsule_sabour2017dynamic} are represented by a vector which contains the features of an object and its likelihood. Higher capsules are activated only when the group of objects below them is consistent in their orientation and size with each other.  The authors tested CapsNet on affNIST dataset to show its robustness to affine transformations. CapsNet was trained on MNIST digits only with random translations and achieved 79\% accuracy on affNIST which confirms it generalizes well to small affine transformations.

\subsection{Deep RL for computer vision}
Deep RL methods have been making steady progress in games \cite{101mnih2013playing}, robotics \cite{102levine2016end}, finance \cite{103deng2016deep}, etc., and have recently been expanded to a wide range of computer vision tasks. \cite{201caicedo2015active} addresses the problem of detection by learning localization policy to find region proposals which best focus on the object while \cite{202kong2017collaborative} suggests joint agent detection to reduce the iterations compared to a single agent. 

For visual tracking, \cite{203yun2017action} proposes to dynamically track the object with increased efficiency in search space through selecting sequential actions on candidate bounding boxes. \cite{204ren2018collaborative} extends to multi-object tracking by modeling each object as an agent and exploiting the collaborative interactions between agents. 

Video recognition can be computationally expensive if performing exhaustive search in every frame. For the first time, \cite{205rao2017attention} authors produce temporal-spatial representations then find the most relevant information in video pairs as a Markov decision process (MDP) for face recognition. In action recognition, \cite{206tang2018deep} propose to reduce the computational burden by selecting only the key frames in skeleton-based videos, and generating the reward with the graph-based convolutional NN.

Deep RL has also been applied to image editing and color enhancement. \cite{207li2018a2} deploys a RL-based method for automatic image cropping where the agent makes sequential decisions on where to crop the original image to maximize the aesthetics score. \cite{208yu2018crafting} progressively restores a corrupted image by selecting a specific tool from the toolbox at each step while \cite{209park2018distort} focuses on automatic color enhancement where the agent takes an interpretable action sequence to produce a retouched image like a human expert. Here, we present a novel RL-based strategy for geometric invariance and this is, to the best of our knowledge, the first work to apply deep RL to its kind.


\begin{figure*}[t]
\centering
\includegraphics[width=0.83\textwidth]{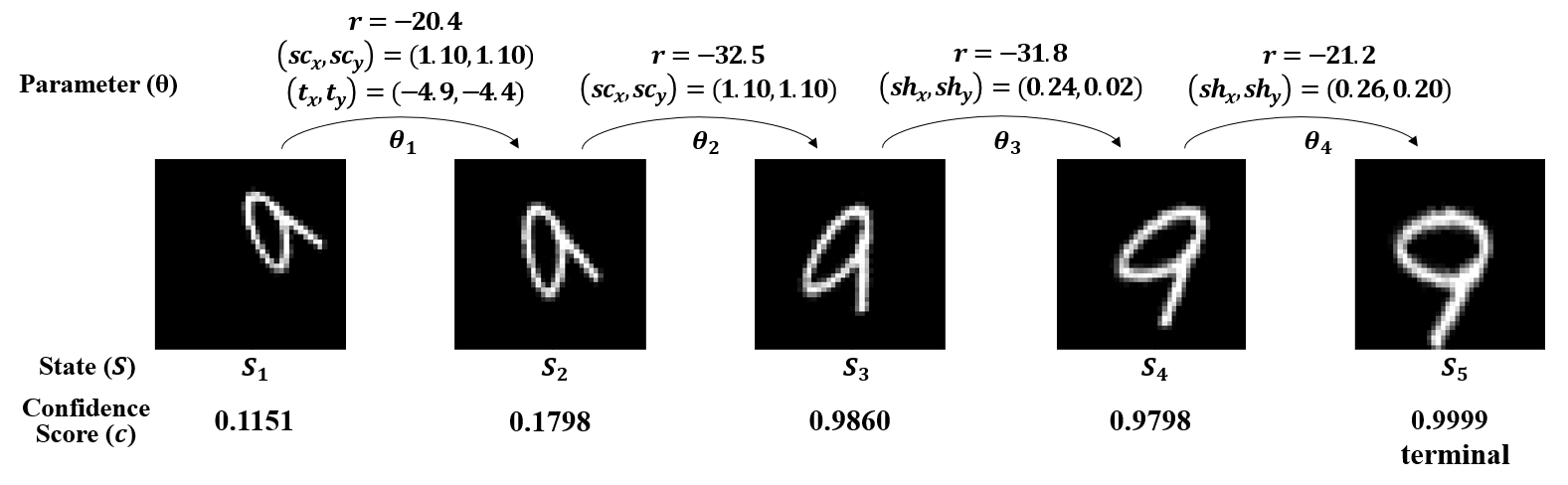} 
\caption{The sequence of transformation of an input image. A distorted input image gets a sequence of parameter to transform it into a canonical style. The sequence terminates when the transformed image gives a high confidence score. It is interesting to see that the final image resembles a style of typical MNIST data.}
\label{fig2}
\end{figure*}


\section{Method}

In this section, we describe the procedure of REST. The purpose of REST is to warp the input image for reliable prediction of the black-box model. We first define the black-box model which performs a classification task and provides the confidence scores. Our method, REST, is composed of two modules. The first part is a trainable RL module which outputs the parameters of the transformation when geometrically deformed images are given as states. The other is a warp module which takes an input image and performs warping with the transformation parameters.

\subsection{Black Box Model}

We view the black-box function as a probabilistic model $P\left(y|x\right)$ where the model assigns a probability of the class $y$ for an input $x$. For the black-box image classifier $B$, $x \in \mathbb{R}^{H \times W \times C}$ is an input image, where $H$, $W$, $C$ are the height, width, and channels of an input image, and $y \in \mathcal{Y}$ is the class with $\mathcal{Y}$ a set of classes. We formulate the function of black-box as $y$ and $c = B\left(x\right)$, where $c$ is the confidence score obtained from the black box $B$.  We discuss the definition of confidence score and which method we choose to measure it in the next section. The classifier $B$ is trained on the dataset $D = \left\{\left(x_i, y_i\right) | 1 \leq i \leq  n, \; i \in \mathbb{N} \right\}$, and $x_i$ is drawn from the data distribution $\mathcal{P}\left(x\right)$ which is regarded as in-distribution. When training the REST learner, we are not allowed to access the original dataset $D$ to ensure black-box assumption.

\subsection{Confidence Score}

It is important for neural networks to understand how uncertain they are in their predictions. While there are methods to quantify the predictive uncertainty by training neural networks where the structure is different from that of prevailing deterministic models \cite{graves2011practical,blundell2015weight}, there are also other approaches to obtain the uncertainty without modifying the existing classifier structure or training procedure \cite{gal2016uncertainty,gal2016dropout,ritter2018a}. We choose one of the latter methods as we are dealing with the black-box model. We here set a confidence score $c$ to be inversely proportional to the predictive uncertainty obtained by the black-box model. 

One of the most widely used approaches to uncertainty estimation is Monte-Carlo (MC) Dropout due to its simplicity \cite{gal2016uncertainty,gal2016dropout}. When the black-box classifier has a dropout layer, the uncertainty is calculated by performing dropout at the inference time. Another simple method for measuring uncertainty is using a value proportional to the inverse of the prediction probability $\mathcal{P}\left(y|x\right)$ (which means the prediction probability is a confidence score). 

While there exists an extensive literature that softmax outputs are sometimes overly confident and thus not sufficient to consider it as confidence scores \cite{gal2016uncertainty,gal2016dropout}, in our experiment, the output of the softmax itself was a proper indicator for confidence level of predictions. Experimental results of replacing a confidence score with predictive uncertainty by MC dropout are given in Appendix B.  

\subsection{Reinforcement Spatial Transform Learner}

REST learner consists of two modules: the RL module and the warp module. Given a distorted input image $x^\prime$, the RL module provides $\theta$ which is used for the warp module $T_\theta$ to transform a distorted image $x^\prime$ into $x^\prime_{tf}$. As in-distribution $\mathcal{P}\left(x\right)$ and the dataset $D$ are unknown, we indirectly estimate the possibility of $x^\prime_{tf}$ by the confidence score from in-distribution sample. 

we train the RL module with a new dataset $D^\prime = \left\{\left(x^\prime_i, y^\prime_i\right) | 1 \leq i \leq  n, \; i \in \mathbb{N} \right\}$ to find a transform parameter $\theta$ for each data $x^\prime_i$ that makes a higher confidence score for the transformed image $T_\theta\left(x^\prime_i\right)$. Let us clarify the state, action, policy, and reward of the RL module. State $S_t$ is an input data in time step $t$ with $S_1 = x^\prime_i$, and action $A_t$ is a parameter $\theta$ used in the warp module $T_\theta$. By mapping $S_t$ by $T_\theta$, we get a next state $S_{t+1} = T_\theta\left(S_t\right)$. Policy $\pi\left(A_t | S_t\right)$ is a probability for each parameter $\theta$ to be selected given input state, where the purpose of training the RL module is to find the optimal policy. 

As the agent is trained to maximize a cumulative reward, we formulate a Reward $R_t$ to increase as the confidence score $c_{t+1}$ of the image $S_{t+1}$ is bigger than $c_t$. 

\begin{equation}
R_t = \log\left(c_{t+1}\right) - \log\left(c_t\right)
\end{equation}
\\
In training mode, we set the confidence score to be the the prediction probability of the target class label, $c_t = \mathcal{P}\left(y^\prime_i | S_t\right)$ when $S_1 = x^\prime_i$. Then the equation (1) can be interpreted as a difference between log-likelihood of the target class. 

To get higher reward rate when the confidence score gets closer to 1, and to give penalty per step for a shorter length of episode, we modify the reward function as 

\begin{equation}
    R_t = \left(-\log\left(1-c_{t+1}\right)\right) - \left(-\log\left(1-c_t\right)\right) - 1
\end{equation}
\\We also perform an ablation study that compares reward function (1) and (2) in Appendix C.
In training mode, episode terminates when the confidence score $c_{t+1}$ exceeds a threshold value in time step $t$ or the number of steps reaches max number. We set a max number to be 10 for all the experiments. In inference mode, we use maximum likelihood, $max_{y \in \mathcal{Y}} \mathcal{P}(y|S_{t+1})$, instead of the prediction probability of the target class, $\mathcal{P}\left(y^\prime_i | S_t\right)$, for the confidence score to determine the termination, and the other parts are the same. 

\subsection{Training}

We use PPO algorithm \cite{schulman2017proximal} for the RL module. To make it operate in continuous action space, we modify the actor network where the output is a mean of the Normal distribution with standard deviation to be 1. The re-parameterization trick is used to enable backpropagation. After setting the activation function of actor network to the hyperbolic tangent function, we change the output bounds of $\left(-1, 1\right)$ to match action bounds so that the parameter $\theta$ exists within action bounds, making stable learning. In the test phase, we reduce the standard deviation of a normal distribution to decrease the randomness of actor network. We train the model using Adam optimization \cite{adam2015optimization} with the learning rate 0.0001 and batch size 256.


\begin{table}[t]
\centering
\resizebox{.95\columnwidth}{!}{
\smallskip\begin{tabular}{c||ccccc}
Method  & R     & RSc   & RSh   & RSS   & RSST  \\ \hline
BB      & 69.28 & 65.52 & 56.52 & 53.95 & 17.99 \\
REST+BB & \textbf{97.63} & \textbf{96.30} & \textbf{94.81} & \textbf{93.20} & \textbf{85.05}
\end{tabular}
}
\caption{The accuracy of a distorted MNIST dataset. BB refers to a black-box model. While it presents a low performance when using only a black-box model, by applying the REST, the accuracy increases significantly. }\smallskip
\label{table1}
\end{table}


\begin{table*}[t]
\centering
\resizebox{.7\textwidth}{!}{
\smallskip\begin{tabular}{cc||cccccc}
Dataset & Method                & base  & R              & RSc            & RSh            & RSS            & RSST           \\ \hline
\multirow{2}{*}{SVHN}    & BB        & 96.03 & 59.96          & 56.46          & 57.05          & 53.54          & 26.09          \\
                         & REST + BB & -     & \textbf{89.38} & \textbf{88.58} & \textbf{89.00} & \textbf{85.92} & \textbf{83.82} \\ \cline{1-2}
\multirow{2}{*}{CIFAR10} & BB        & 93.78 & 51.58          & 49.14          & 50.69          & 48.16          & 32.09          \\
                         & REST + BB & -     & \textbf{74.33} & \textbf{72.15} & \textbf{70.73} & \textbf{69.46} & \textbf{60.27} \\ \cline{1-2}
\multirow{2}{*}{STL10}   & BB        & 77.59 & 41.57          & 38.46          & 41.26          & 37.97          & 30.90          \\
                         & REST + BB & -     & \textbf{62.24} & \textbf{59.18} & \textbf{59.94} & \textbf{56.55} & \textbf{53.20}
\end{tabular}
}
\caption{ The accuracy of a distorted real-world dataset. We show that applying our model also performs better in real-world datasets such as SVHN, CIFAR10, and STL10. }\smallskip
\label{table2}
\end{table*}


\begin{figure}[t]
\centering
\includegraphics[width=0.8\columnwidth]{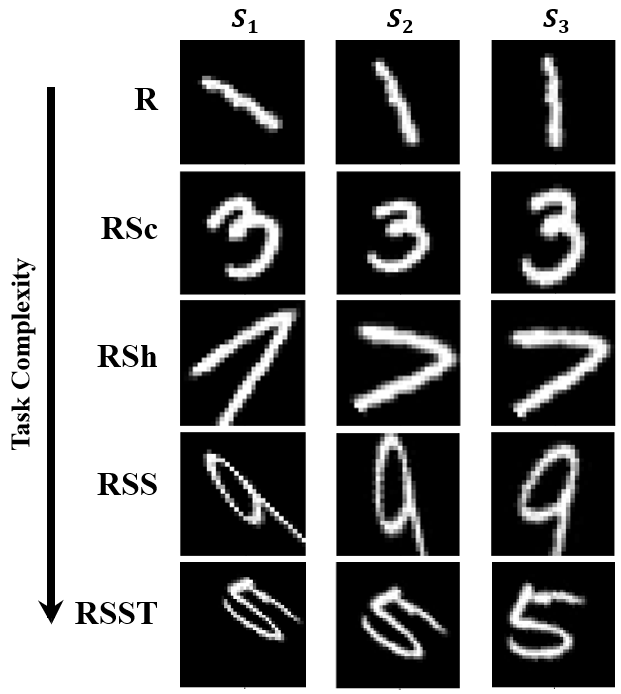} 
\caption{Various transformations of different complexity. $S_1$ is the initial state(a deformed image) and sequence of best actions are taken given the states.}
\label{fig3}
\end{figure}


\section{Experiment}

In this section we demonstrate the effectiveness of our approach in generalization by attaching the REST learner to the black-box model. To train the RL module, we generate a dataset $D^{\prime}$ by applying a random affine transformation to the expected canonical style of the dataset $D$ used for training the black box classifier. For example, if we know that a black-box model $B$ classifies a gray-scale numeric image, it is plausible to assume that the model $B$ is trained by the MNIST dataset, and therefore we generate a dataset $D^\prime$ by random affine transformation of MNIST images.  

We also use the affine transformation for the warp module $T_\theta$. Although 6 parameters are typically used for affine matrix, we factorize the affine matrix into 4 matrices to have 7 parameters, $\theta = \left(\theta_{r}, \theta_{sc1}, \theta_{sc2}, \theta_{sh1}, \theta_{sh2}, \theta_{t1}, \theta_{t2}\right)$.

\begin{equation}
\begin{split}
    T_{\theta}
    = & 
    \begin{bmatrix}
      \cos\theta_r & -\sin\theta_r & 0 \\
      \sin\theta_r & \cos\theta_r & 0 \\ 
      0 & 0 & 1 \\ 
    \end{bmatrix}
    \cdot
    \begin{bmatrix}
      \theta_{sc1} & 0 & 0 \\
      0 & \theta_{sc2} & 0 \\ 
      0 & 0 & 1 \\ 
    \end{bmatrix}
    \\
    \cdot & 
    \begin{bmatrix}
      1 & \theta_{sh1} & 0 \\
      \theta_{sh2} & 1 & 0 \\ 
      0 & 0 & 1 \\ 
    \end{bmatrix}
    \cdot
    \begin{bmatrix}
      1 & 0 & \theta_{t1} \\
      0 & 1 & \theta_{t2} \\ 
      0 & 0 & 1 \\ 
    \end{bmatrix}
\end{split}
\label{eqt:numerical_scheme1}
\end{equation}
\\
where $\theta_{r}$, $\theta_{sc}$, $\theta_{sh}$, and $\theta_{t}$ are parameters for rotation, scaling, shearing, and translation, respectively. Details of generating affine-warped dataset and settings of action bounds are described in Appendix A.

As we use the image dataset for experiments, the actor and the critic networks in RL module are constructed by two convolution layers followed by two fully connected layers. We choose 5-layer CNN and STN\cite{STN_jaderberg2015spatial} as baselines and test the transformation invariance on affine-warped MNIST, SVHN, CIFAR10, and STL10 datasets. We train the baseline models using Adam optimization\cite{adam2015optimization} with the learning rate 0.0001 and batch size 128.

We start with the distorted MNIST dataset to show the improvement of classification performance in a shifted data distribution. We further test our model in a more challenging real-world dataset such as SVHN, CIFAR10, and STL10. Then, we demonstrate the ability of generalization of our model in a more shifted, arduous setting where disjointing subsets are selected for training and testing the REST learner. Finally, we show that even with training a REST learner with a small number of training data, the performance does not drop significantly, which results in good sample efficiency.


\begin{figure*}[t]
\centering
\includegraphics[width=0.75\textwidth]{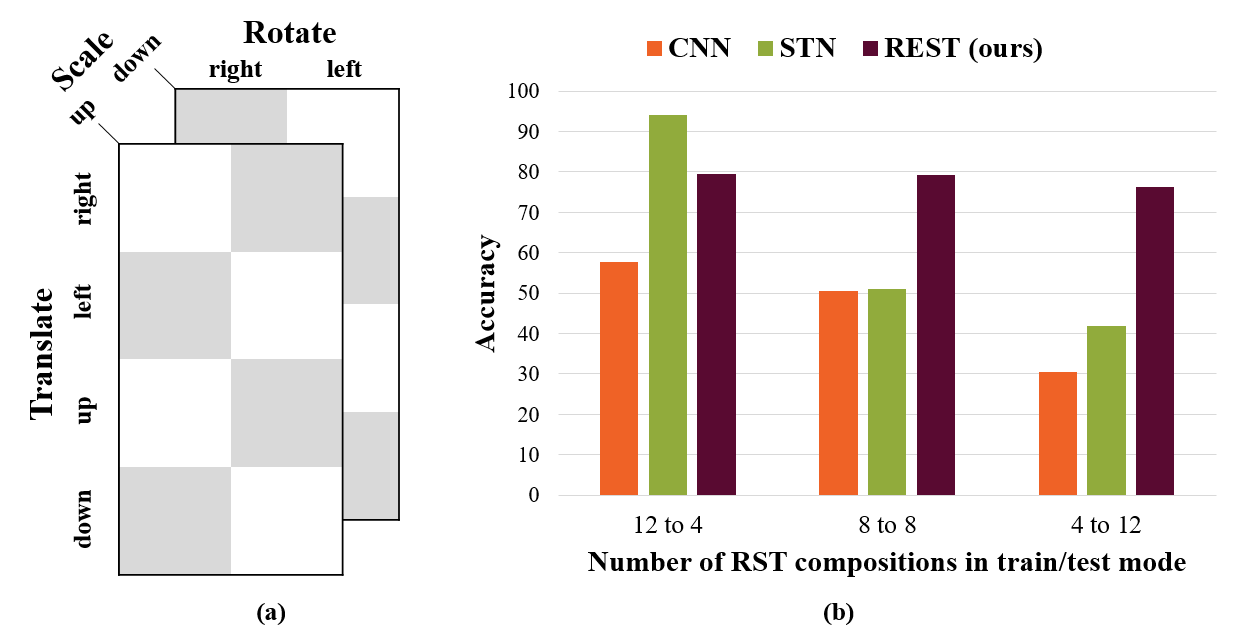} 
\caption{The first figure is an example of separating the composition of RST transformation into two disjoint subsets. The gray parts are used for generating distorted training data while the white parts are for deformed test data. The second figure shows test results on three different decomposition setting. Our models shows the best generalization effect as the task gets harder.}
\label{fig4}
\end{figure*}

\subsection{Improvement of Classification Performance}

To evaluate the improvement of black-box\textquotesingle s performance by using REST learner, we begin with the black-box gray-scale numeric image classifier. We use the CNN pre-trained with MNIST dataset as a black-box classifier which performs 99\% accuracy in MNIST test data. However, when applying randomly rotated MNIST test images (R), the accuracy decreases to 69.28\%. By attaching the REST in front of the black-box model, the accuracy increases back to 97.63\%. 

We also experiment with more difficult tasks by generating the data with rotation and scaling (RSc), rotation and shearing (RSh), rotation, scaling, and shearing (RSS), and rotation, scaling, shearing, and translation (RSST) of the MNIST test digits. 

Examples of the sequence of transformation for each warping method are shown in Figure~\ref{fig3}. All the affinely warped input images in the first column are transformed two times to get
the final output. The quantitative results are shown in Table 1. As task complexity increases from simple rotation to full affine transformation, the black-box classifier gets worse for predicting the label. For the RSST method, the accuracy of the black box model is 17.99 percent, which is almost random choice. However, by adding the REST model in front of the black-box model, it highly improves the performance. The difference in accuracy between applying the REST model and only using black-box model gets larger as the task becomes more difficult.

Figure 2. shows the sequence of how the distorted input image is transformed by REST. When affinely extended image of number 9 is taken for the first state $S_1$, the RL module outputs transform parameter $\theta = \left(20.4, 1.1, 1.1, 0, 0, -4.9, -4.4\right)$. Then, the warp module receives the state $S_1$ and the parameter $\theta$ to produce the next state $S_2$, which is rotated clockwise, scaled up, and translated to the upper-left direction. The black-box model maps the transformed image $S_2$ to a confidence score $c_2$. As the confidence score does not exceed a terminal threshold, the process iterates until the terminal condition is satisfied. An interesting observation is that the final image $S_5$ seems to resemble a style of data sampled from in-distribution $D$, which is the MNIST dataset in this case. The number 9 in the image $S_5$ is standing upright and locating at the center of an image with its size similar to typical MNIST data. 

We further evaluate our model in real-world datasets. We use three datasets which are SVHN, CIFAR10, and STL10. For the black-box models, we use CNN models that perform 96.03\%, 93.78\%, and 77.59\% for accuracy, respectively. Table 2 shows the test accuracy of black-box classifier before and after attaching REST. The performance improves when the REST model is applied at the front-end of the black-box model.


\subsection{Generalization}

In the previous section, we have shown that the REST works well by interacting with the black-box model. However, the limitation of the previous experiment is that the training and test data are generated in the same transformation format. In this section, we demonstrate the effect of generalization by generating the training data and test data in a different condition.
 
We first set the black box model as a general MNIST classifier. We then generate a distorted MNIST dataset by rotation, scaling, and translation (RST) to train the REST learner. We constrain the behavior of transformation by choosing one in (right, left), (up, down), (right, left, up, down) for each transformation R, S, and T, respectively. Therefore, there are 16 possible transformations of RST. We split 16 transformations of RST into two disjoint subsets, and apply each of the subset to MNIST for generating the training data and the test data. Figure~\ref{fig4}(a) shows an example of separating 16 transformations into two subsets, each containing 8 transformations. 

We compared the REST model with CNN and STN, and the results are shown in Figure~\ref{fig4}(b). When the training dataset is generated by 12 RST transformations, STN performs better than ours. However, as the number of RST compositions used for transforming the training data is decreasing, CNN and STN suffer from predicting the correct labels. On the contrary, our model has a slight decrease in accuracy. It means that our model has better generalization in data shift conditions. It is assumed that our model shows a generalization effect because our model takes exploration in the training process, which generates samples that are not present in the training data distribution. 


\begin{table}[t]
\centering
\resizebox{.95\columnwidth}{!}{
\begin{tabular}{c||ccccc}
    Method  & R              & RSc            & RSh            & RSS            & RSST           \\ \hline
CNN   & 87.24          & 82.91          & 76.67          & 70.97          & 38.16          \\
CNN++ & 88.69          & 85.86          & 80.48          & 76.58          & 39.92          \\
STN   & 55.95          & 38.60          & 34.87          & 27.83          & 14.94          \\
STN++ & 93.63          & 92.13          & 89.37          & 85.39          & 80.80          \\ 
REST (ours)  & \textbf{97.02} & \textbf{95.94} & \textbf{93.45} & \textbf{92.79} & \textbf{83.18}
\end{tabular}
}
\caption{The accuracy of models trained by a small number of data. We trained CNN, STN, and REST by 1000 affine-warped training data. CNN++ and STN++ are trained by 56000 data where 55000 is a typical MNIST training data and the rest 1000 is affine-warped training data. The result shows that our model performs best on all different kinds of transformation.}\smallskip
\label{table3}
\end{table}


\subsection{Sample Efficiency}

In this section, we examine the effect of sample efficiency of our model. While 55,000 randomly affine-warped data has been generated from 55,000 training MNIST data to train the REST learner, in this section, we only generate 1000 randomly affine-warped data from 1000 training MNIST data. Then we perform the same process for training and test the model. We compare our model with CNN and STN. Both networks are trained by 1000 data. We also trained CNN and STN with 56,000 data, where 55,000 data is MNIST training data and the other 1000 data is randomly affine-warped data. We call it CNN++ and STN++, respectively. All models are tested with 10,000 affine-warped test data and the results are shown in Table 3.

 As a few data is used for training CNN and STN, they result in low accuracy for all different types of transformations. Also, although many data is used for training CNN++ and STN++, they also show a low performance because only a small number of training data is in the same distribution with that of test data. On the contrary, our model has a best performance. It is considered that our model shows a sample efficiency because of the exploration process in the training step. In the process of creating a state to train the model, new images are created for each episode. These images can be considered as training data which is implicitly augmented.


\section{Conclusion}

Robustness to real-world data is essential for deep learning systems to be successfully deployed in reality. Most studies so far have focused on improving generalization performance on test datasets when a single whole dataset is randomly split into training and test dataset. In other words, the training and test dataset are drawn from the same distribution and have similar sample statistics. Under this experimental assumption, the performance on the real-world data is not necessarily guaranteed even if the system generalizes well on the test dataset. 

In this paper, we addressed geometric invariance using deep RL by transforming out-of-distribution samples into training distribution of the pretrained black box classifier in the system. We showed that the proposed method, REST, can improve the robustness of deep learning systems to various image warping. Specifically, as the complexity of the task gradually increased from simple rotation to full affine transformation, i.e. from one to six degrees of freedom, the relative performance of REST over the black-box model also increased accordingly.

We analyzed the generalization performance on unknown transformations by defining 16 disjoint subsets of affine transformations. REST generalized better as we trained it with a fewer number of  transformation combinations while more of novel and unseen transformations were given at test time. Lastly, we experimented baseline methods with only 1000 affine-warped training data and showed REST is efficient in learning with a small number of samples as well. The action space of our method is focused on geometric transformation in this work but can be extended to other image processing techniques such as auto exposure, white balancing, edge enhancement, noise reduction to fill the gap between controlled experimental settings and real-world scenarios in the future works.

\section{ Acknowledgments}
This work is in part supported by Basic Science Research Program (NRF-2017R1A2B2007102) through NRF funded by MSIT, Technology Innovation Program (10051928) funded by MOTIE, Bio-Mimetic Robot Research Center funded by DAPA (UD190018ID), Samsung Electronics AI Grant,  MSIT-IITP grant (No.2019-0-01367, BabyMind), INMAC, and BK21-plus.


\bibliographystyle{aaai}
\bibliography{main}

\end{document}


\maketitle

\section{Appendix A. Experimental Setting}

We describe details of experimental setting. We first illustrate how to generate the dataset for training and testing the REST learner. Then, we illustrate the hyper-parameter setting of action bounds.

\subsection{Generating Distorted Data}

\subsubsection{classification performance and sample  efficiency} 

For the experiment regarding the classification performance and sample efficiency, we generate dataset in 5 different transform method which is R, RSc, RSh, RSS, and RSST.
First, to make a distorted MNIST dataset by transform method R, we sample a rotation angle $\alpha_r$,

$$\alpha_r \sim \mathcal{U}\left(\mathcal{I}\left[-50,-20\right] \cup \mathcal{I}\left[20,50\right]\right)$$

where $\mathcal{U}$ is a uniform distribution and $\mathcal{I}$ is an interval. Each image in the MNIST dataset is rotated by applying the sampled angle. 
In the same way, scaling value $\alpha_{sc1}$ and $\alpha_{sc2}$, shearing value $\alpha_{sh1}$ and $\alpha_{sh2}$, and translation value $\alpha_{t1}$ and $\alpha_{t2}$ is sampled from uniform distribution,

$$\alpha_{sc1} \sim \mathcal{U}\left(\mathcal{I}\left[0.8,0.9\right] \cup \mathcal{I}\left[1.1,1.2\right]\right)$$
$$\alpha_{sc2} \sim \mathcal{U}\left(\mathcal{I}\left[0.8,0.9\right] \cup \mathcal{I}\left[1.1,1.2\right]\right)$$
$$\alpha_{sh1} \sim \mathcal{U}\left(\mathcal{I}\left[-0.5,-0.2\right] \cup \mathcal{I}\left[0.2,0.5\right]\right)$$
$$\alpha_{sh2} \sim \mathcal{U}\left(\mathcal{I}\left[-0.5,-0.2\right] \cup \mathcal{I}\left[0.2,0.5\right]\right)$$
$$\alpha_{t1} \sim \mathcal{U}\left(\mathcal{I}\left[-6,-3\right] \cup \mathcal{I}\left[3,6\right]\right)$$
$$\alpha_{t2} \sim \mathcal{U}\left(\mathcal{I}\left[-6,-3\right] \cup \mathcal{I}\left[3,6\right]\right)$$

$\left(\alpha_{r}, \alpha_{sc1}, \alpha_{sc2}\right)$, $\left(\alpha_{r}, \alpha_{sh1}, \alpha_{sh2}\right)$, $(\alpha_{r}, \alpha_{sc1}, \alpha_{sc2}, \alpha_{sh1},$ $\alpha_{sh2})$, and $\left(\alpha_{r},\alpha_{sc1},\alpha_{sc2},\alpha_{sh1},\alpha_{sh2},\alpha_{t1},\alpha_{t2}\right)$ is used in each method RSc, RSh, RSS, and RSST for generating distorted MNIST dataset. We exclude intermediate values from the sampling range to ensure that the typical form of MNIST is not included in the distorted MNIST set. We use the same sampling range for generating transformed real-world datasets.

\subsubsection{generalization}

For the experiment regarding generalization, we use the RST transformation method where R, S, and T is rotation, scaling, and translation, respectively. We increase the gap of the interval range of the uniform distribution to make more distorted images compare to the generating method above. 

$$\alpha_r \sim \mathcal{U}\left(\mathcal{I}\left[-60,-50\right] \cup \mathcal{I}\left[50,60\right]\right)$$
$$\alpha_{sc1} \sim \mathcal{U}\left(\mathcal{I}\left[0.75,0.8\right] \cup \mathcal{I}\left[1.2,1.25\right]\right)$$
$$\alpha_{sc2} \sim \mathcal{U}\left(\mathcal{I}\left[0.75,0.8\right] \cup \mathcal{I}\left[1.2,1.25\right]\right)$$
$$\alpha_{t1} \sim \mathcal{U}\left(\mathcal{I}\left[-7,-6\right] \cup \mathcal{I}\left[6,7\right]\right)$$
$$\alpha_{t2} \sim \mathcal{U}\left(\mathcal{I}\left[-7,-6\right] \cup \mathcal{I}\left[6,7\right]\right)$$

\subsection{Action Bounds}

The actor network in RL module prints transform parameter $\theta = \left(\theta_{r}, \theta_{sc1}, \theta_{sc2}, \theta_{sh1}, \theta_{sh2}, \theta_{t1}, \theta_{t2}\right)$ as an output. As we consider a continuous action space, we set the action bounds of each elements in parameter $\theta$ to have stable learning. We set the action bounds as follows.

$$ \theta_{r} \in \left[-30,30\right] $$
$$ \theta_{sc1} \in \left[0.9,1.1\right] $$
$$ \theta_{sc2} \in \left[0.9,1.1\right] $$
$$ \theta_{sh1} \in \left[-0.2,0.2\right] $$
$$ \theta_{sh2} \in \left[-0.2,0.2\right] $$
$$ \theta_{t1} \in \left[-4,4\right] $$
$$ \theta_{t2} \in \left[-4,4\right] $$

\section{Appendix B. Confidence Score by MC Dropout}

In case of severe geometric distortion, we assume the distorted images are off-manifold of normal images. We thus experiment using Mutual Information (MI) as a confidence score since MI is known to be able to detect out-of-distribution images and adversarial examples \cite{smith2018understanding}. 
Mutual Information (MI) between a label $y$ and the model parameters $\theta$ assesses the inconsistency between multiple models which is regarded as a measure of \emph{epistemic uncertainty} \cite{gal2016uncertainty,gal2016dropout}.\\

The MI can be expressed as
\begin{align}\label{equ_MI}
I(y, \theta \mid x, D)&=H[p(y\mid x,D)] - \mathbb{E}_{p(\theta|D)}\big[ H\left[ y\mid x,\theta \right] \big]  
\end{align}
The first and second terms on the RHS of \eqref{equ_MI} are interpreted as \emph{predictive entropy} and \emph{aleatoric uncertainty}, respectively. We compute these quantities using MC estimation:
\begin{align}
I(y, \theta \mid x, D)&\approx H[\frac{1}{T}\sum_{i=1}^{T}p(y\mid x,\theta_i)] - \frac{1}{T}\sum_{i=1}^{T}H[p(y\mid x,\theta_i)]  
\end{align}

We experiment with MI instead of prediction probability as the confidence score and evaluate the epistemic uncertainty with 30 Monte Carlo samples.

\begin{table}[h]
\centering
\resizebox{.95\columnwidth}{!}{
\smallskip\begin{tabular}{c||ccccc||c}
Method    & R     & RSc   & RSh   & RSS   & RSST  & Reward    \\ \hline
BB        & 69.28 & 65.52 & 56.52 & 53.95 & 17.99 & -         \\
REST + BB & 95.52 & 90.31 & 91.02 & 87.99 & 11.58 & not tuned \\
REST + BB & 97.32 & 96.62 & 95.45 & 95.15 & 80.37 & tuned    
\end{tabular}
}
\caption{Performance comparison with MI as confidence score}\smallskip
\label{sup_table1}
\end{table}

In training mode, we set the confidence score to be the mutual information of the target class, $c_t = I(y, \theta \mid S_i, D)$ where $S_i = x_i ’$. To get a higher reward rate when the confidence score gets closer to 0, we modify the reward function as
$R_t = (-\log c_{t + 1}) - (-\log c_t)-1$. The experiment was conducted on random affine transformed MNIST images with the same settings except for the reward function.

Experiment result shows good performance in R, Rsc, Rsh and RSS with 95.52\%, 90.31\%, 91.02\%, and 87.99\% accuracy, respectively. However, the performance was lower than when using the reward function based on prediction probability. In RSST, the accuracy is 11.58\%, lower than the 17.99\% of accuracy when only using the black-box classifier, which means REST is not learning at all.


\begin{figure}[t]
\centering
\includegraphics[width=0.8\columnwidth]{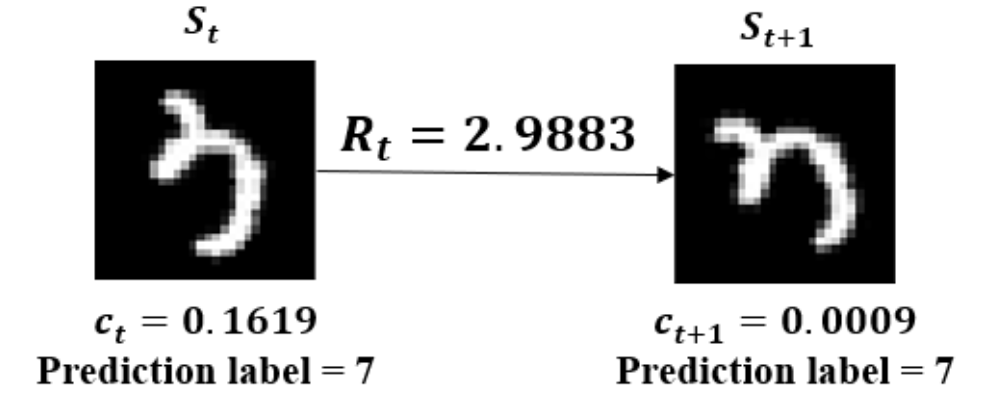} 
\caption{Incorrect rewards can be provided when black-box classifier makes overconfident predictions on OOD examples.}

\label{fig_appendix_MI}
\end{figure}

The reason for poor learning is that lowering the confidence score, that is, uncertainty, sometimes does not improve performance. If you look at Figure \ref{fig_appendix_MI}, agent should be penalized because it rotated number 3 image $S_t$ in the wrong direction. However, reward $R_t$ became positive due to lower uncertainty of seeing $S_2$ as number 7, which is because DNN in black box classifier makes overconfident yet incorrect predictions on this out-of-distribution sample. If the uncertainty for the wrong prediction is lowered, reward should be negative as a penalty, but it is rather positive with $R_t$ set above.

We decided to tune the reward function to overcome this problem. In setting the reward function, the prediction label as well as the uncertainty of the black box classifier are taken into account. Dividing all cases due to changes in uncertainty and prediction labels, the reward is defined as $R_t = (-\log c_{t + 1})-(-\log c_t)-1$ or $R_t = (-\log c_t)-(-\log c_{t+1})-1$ as appropriate to be positive or negative.

Experiment result shows that the performance improves in R, RSc, RSh, and RSS, showing 97.32\%, 96.62\%, 95.45\%, and 95.15\% accuracy, respectively, with the same or better performance than using the reward function based on prediction probability. In RSST, however, the performance decreased slightly to 80.37\% compared to using prediction probability as reward function. The experiment results are summarized in Table \ref{sup_table1}.

Using uncertainty instead of prediction probability to define reward function is an unsatisfactory experiment because we need to tune the reward function for good performance. This is because MI we used did not completely solve the problem of overconfidence of the neural network to out-of-distribution data. If the uncertainty that solves this problem is used for confidence score, even better performance can be achieved than with the prediction probability.

\section{Appendix C. Ablation Study for Reward Design}

We experiment an ablation study for reward design of equation (1) and (2). Experiments results below is the mean of five repeated experiments.

\begin{table}[h]
\centering
\resizebox{\columnwidth}{!}{%
\smallskip\begin{tabular}{c||c|c||ccccc}
Reward        &  & Phase & R       & RSc   & RSh   & RSS   & RSST      \\ \hline
\multirow{5}{*}{(1)}    & \multirow{2}{*}{time(sec)}         & Train     & 419   & 461   & 579    & 591   & 1,547 \\
                                                  &                           &Test       & 31.3 & 32.7 & 35.1 & 36.3 & 156.8 \\ \cline{2-3}
                                                  & \multirow{2}{*}{avg seq length} & Train & 1.89 & 2.12 & 2.96 & 3.08 & 8.53 \\
                                                  &                           & Test    & 1.36 & 1.37 & 1.52 & 1.56 & 7.91 \\ \cline{2-3}
                                                  & test accuracy   &            & 97.6 & 96.2 & 94.9 & 94.2 & 35.65\\\hline
\multirow{5}{*}{(2)}    & \multirow{2}{*}{time(sec)}         & Train     & 419   & 443   & 561    & 562   & 1,285 \\
                                                  &                           &Test       & 30.7 & 30.7 & 32.9 & 33.1 & 64.2 \\ \cline{2-3}
                                                  & \multirow{2}{*}{avg seq length} & Train & 1.89 & 2.04 & 2.89 & 2.90 & 6.86 \\
                                                  &                           & Test    & 1.34 & 1.31 & 1.46 & 1.43 & 3.00 \\ \cline{2-3}
                                                  & test accuracy   &            & 97.6 & 96.2 & 95.7 & 93.4 & 81.69\\
\end{tabular}%
}
\caption{Performance comparison of reward function (1) and (2)}\smallskip
\label{sup_table2}
\end{table}

In the experiments, we compare reward functions (1) and (2) in terms of training and testing time, average sequence length, and test accuracy. Table \ref{sup_table2} shows that the performance of the two reward functions do not differ much on relatively easier tasks such as simple rotation, but reward function (2) performs significantly better on more complex tasks such as RSST.

\bibliographystyle{aaai}
\bibliography{main}